\pdfoutput=1

\documentclass[11pt,a4paper]{article}

\input{minted_cache/default-pyg-prefix.pygstyle} %
\input{minted_cache/default.pygstyle} %

\usepackage[hyperref]{emnlp2020}
\usepackage{times}
\usepackage{latexsym}

\usepackage{booktabs}
\usepackage{xspace}
\usepackage{graphicx}
\usepackage{pgfplots}
\usepackage{amsmath}
\usepackage{todonotes}
\usepackage{fancyvrb}
\usepackage[utf8]{inputenc}
\usepackage{xcolor}

\usepackage{algorithm,algorithmicx}
\usepackage[noend]{algpseudocode}

\usepackage{enumitem}

\usetikzlibrary{pgfplots.groupplots}

\DeclareMathOperator*{\argmax}{\arg\!\max}

\DeclareMathOperator*{\embed}{{\cal{M}}}

\usepackage{microtype}

\usepackage{subcaption}
\usepackage{adjustbox}

\aclfinalcopy %
\def\aclpaperid{16} %

\def\Snospace~{\S{}}

\title{Paraphrase Generation as Zero-Shot Multilingual Translation:\\ Disentangling Semantic Similarity from Lexical and Syntactic Diversity}

\author{Brian Thompson \\
  Johns Hopkins University \\
  \texttt{brian.thompson@jhu.edu} \\\And
  Matt Post \\
  Johns Hopkins University \\
  \texttt{post@cs.jhu.edu} \\}

\date{}

\begin{document}
\maketitle

\begin{abstract}

Recent work has shown that a multilingual neural machine translation (NMT) model
can be used to judge how well a sentence paraphrases another sentence in the same language \cite{prism};
however, attempting to \emph{generate} paraphrases
from such a model using standard beam search produces trivial copies or near copies.
We introduce a simple paraphrase generation algorithm 
which discourages the production of n-grams that are present in the input.
Our approach enables paraphrase generation in many languages from a single multilingual NMT model.
Furthermore, the amount of lexical diversity between the input and output can be controlled at generation time.
We conduct a human evaluation to compare our method to a paraphraser trained 
on the large English synthetic paraphrase database
ParaBank 2 \cite{hu-etal-2019-large}
and find that our method produces paraphrases that better preserve meaning
and are more gramatical, 
for the same level of lexical diversity.
Additional smaller human assessments demonstrate our approach also works in two non-English languages.

\end{abstract}

\section{Introduction}\label{intro}

Paraphrase generation is the task of
producing a fluent output sentence which is semantically similar
to the input sentence
while being syntactically and/or lexically different from it \cite{bhagat-hovy-2013-squibs}.
Paraphrasing has been of longstanding interest in the NLP community \cite{mckeown-1983-paraphrasing}
and has been used for data augmentation in
question answering \cite{dong-etal-2017-learning, gan-ng-2019-improving},
machine translation (MT) \cite{ hu-etal-2019-improved, khayrallah-etal-2020-simulated},
task oriented dialog \cite{niu-bansal-2018-adversarial, niu-bansal-2019-automatically},
and MT metrics \cite{banerjee-lavie-2005-meteor, zhou-etal-2006-paraeval, denkowski-lavie-2010-extending, prism}.

\citet{prism}
recently released the Prism MT metric, 
which uses a multilingual neural MT (NMT) model as a paraphraser 
to \emph{score} paraphrastic pairs;
they treat paraphrasing as a zero-shot translation task (e.g., ``translation'' from English to English)
and force-decode and score MT system outputs
conditioned on their respective human translations.
They denote their paraphraser as \emph{lexically/syntactically unbiased}
as it does \emph{not} prefer output that differs lexically or syntactically from the input;
this is advantageous for an MT metric as it
assigns the highest score to an MT output which matches or nearly matches a human reference,
but generating from the Prism model using standard beam search produces trivial copies or near copies.

We introduce a simple method to enable paraphrase generation from a multilingual
NMT model.\footnote{We release our code at \url{https://github.com/thompsonb/prism}}
Our method discourages the model from producing n-grams that match n-grams in the input sentence.
This serves to lexically bias the output away from the input sentence, resulting in non-trivial paraphrases.

When considered together with Prism model of \citet{prism},
our paraphrase generation approach offers several potential advantages over
the common technique of training a paraphrase model on synthetic paraphrases generated by translating one side of bitext
into the language of the other side
\cite{wieting-etal-2017-learning, wieting-gimpel-2018-paranmt, hu-etal-2019-large}:
\begin{itemize}[topsep=2pt,itemsep=2pt,partopsep=1pt,parsep=1pt]
    \item The fluency/semantic similarity vs lexical diversity trade-off can be controlled at generation time.
    \item The approach works in many languages, with a single model.
    \item The approach addresses an inherent shortcoming in creating synthetic paraphrases from bitext in which ambiguities in one language can create errorful synthetic paraphrases in the other (see \autoref{discussion}).
\item Separating the fluency and semantic similarity model from the lexical and/or syntactic diversity model allows
them to be developed and evaluated with less interdependencies. 
\end{itemize}

We conduct human evaluations to compare our
proposed method
to a strong English baseline paraphraser trained on the ParaBank 2 dataset \cite{hu-etal-2019-large}, which consists of 50 million synthetic examples generated by translating the Czech side of Czech--English bitext into English and pairing it with the original English. 
We find that our method outperforms this baseline---both in terms of semantic similarity and grammaticality---when our system is adjusted to match the lexical diversity of the baseline.
We also present small scale evaluations that suggest our method is effective in other languages.

\section{Related Work}

\paragraph{Paraphrase Generation}
Machine translation techniques can be used to train paraphrase models \cite{quirk-etal-2004-monolingual}.
Another method to generate a paraphrase is to translate
a text to a different language and then back again \cite{mallinson-etal-2017-paraphrasing}. 
Multiple pivot languages can be used 
to lessen the effect of inherent ambiguities
\cite{aziz-specia-2013-multilingual}, at the expense of complication.
Several works have focused on training on paraphrase data, 
including synthetic data created by starting with bitext and translating one side into the language of the other side
to create synthetic paraphrases %
\cite{wieting-etal-2017-learning, wieting-gimpel-2018-paranmt, hu-etal-2019-large}.
Ideas such as adversarial training \cite{iyyer-etal-2018-adversarial}, reinforcement learning \cite{li-etal-2018-paraphrase}, and variational autoencoders \cite{AAAI1816353,chen-etal-2019-multi} have also been explored in the context of paraphrase generation.

\paragraph{Diversity in Generation} 
Creating paraphrases which differ from their input in non-trivial ways is a challenging problem.
\citet{hu-etal-2019-large} used constrained decoding \cite{hokamp-liu-2017-lexically} 
in conjunction with a set of constraints (e.g., avoiding certain words which are present in the input) 
when creating synthetic paraphrases from bitext.
\citet{kajiwara-2019-negative} also used hard constraints, but at decoding time.
Our work is similar but uses ``soft'' constraints (i.e., down-weighting tokens which complete n-grams in the input, but not disallowing them all together).
Another approach is to control generation with syntactic examples \citep{iyyer-etal-2018-adversarial, chen-etal-2019-controllable}
or codes \citep{shu-etal-2019-generating}.

\paragraph{Multilingual NMT} 

\begin{algorithm*}[ht]
    \caption{Before paraphrasing a sentence, \texttt{buildPenalties()} is called to construct a mapping of word prefixes to subwords that require penalties. Then, \texttt{penalize()} is called to modify the model prediction \texttt{targetLogProbs} at every decoder timestep.
    }
    \label{algo}

\input{minted_cache/11CDC518383705DDE613FFC0E241409512AA82312C1119922434E0B4AB7CAB88.pygtex}  %

\end{algorithm*}

Multilingual NMT \cite{dong-etal-2015-multi} 
has been shown to enable zero-shot translation---that is, translation between languages pairs not included in training 
(e.g., translating from Spanish$\rightarrow$Arabic at test time when the model was trained on Spanish$\rightarrow$English and English$\rightarrow$Arabic, but not Spanish$\rightarrow$Arabic)
\cite{johnson-etal-2017-googles, gu-etal-2018-universal, pham-etal-2019-improving}.
\citet{zhou-etal-2019-paraphrases} also explored incorporating paraphrase data into training to improve multilingual NMT performance.

\citet{tiedemann-scherrer-2019-measuring} explored using paraphrase recognition 
to test the semantic abstraction of a fairly small multilingual NMT system trained on Bibles 
and also demonstrate the model's ability to paraphrase in English. 
However, they did not perform a human evaluation of paraphrase quality, 
and \citet{prism} found that simply generating via beam search from a multilingual NMT
model trained on a large general domain corpus results in trivial copies most of the time. 
We build upon \citet{tiedemann-scherrer-2019-measuring} by using a larger, general domain model, 
introducing a novel generation algorithm to produce output with lexical diversity, 
and performing human evaluations.

\paragraph{Paraphrastic similarity}
Similarity between intermediate representations produced by multilingual NMT encoders 
has been used to measure semantic similarity and/or 
paraphrastic similarity \cite{schwenk-douze-2017-learning, wieting-etal-2019-simple, raganato-etal-2019-evaluation}. 
Similarly, Prism \cite{prism} use a multilingual NMT model as a
lexically/syntactically unbiased paraphraser for scoring MT system outputs conditioned on their associated human reference translations. 
We build on this by introducing a lexical bias away from the input at generation time, 
enabling the use of a multilingual NMT model as a generative paraphraser.

\section{Method}
\label{section:method}

Let $x$ and $y$ be sentences, let $\embed(x)$ represent the meaning of $x$, and let $S(x,y)$ measure the lexical and/or syntactic similarity between the two sentences.
Formally, we can state the problem of paraphrase generation as finding $\hat{y}$:
\begin{equation}
 \hat{y} = \argmax_y \left[ p(y\:{\mid}\:\embed(x)) - \alpha S(x,y) \right]
 \label{paraphrase-eq}
\end{equation}
where $\alpha$ controls the semantic similarity and fluency vs lexical and/or syntactic diversity trade-off.

\subsection{Lexically/Syntactically Unbiased Paraphraser}

The intralingual probability $p(y\:{\mid}\:\embed(x))$ can be viewed as a lexically/syntactically unbiased paraphraser.
This model is responsible for producing output which is both semantically similar to the input and fluent,
but has no notion of lexical and/or syntactic diversity.
We use the multilingual NMT system released with Prism to model $p(y\:{\mid}\:\embed(x))$.

\subsection{Lexical Bias}

We choose n-gram overlap as our measure of lexical and/or syntactic similarity $S(x,y)$,
and  propose a simple n-gram overlap measure that penalizes the production of n-grams matching n-grams in the input sequence
to enable the paraphrase generation.
Our proposed algorithm begins by constructing a set of all (word) $n$-grams, $1~\leq~n~\leq~4$, from the input.\footnote{In this work, we assume words are separated by whitespace.
For languages which do not denote word boundaries, our method could likely be applied after tokenizing the input, or by simply treating each SentencePiece token as a word.}
At each decoding step, the algorithm checks whether any of the target vocabulary subwords \emph{begin} the last word of an input $n$-gram.%
\footnote{We apply the penalty at the start of the generation of the last word of an input n-gram so that the decoder is not encouraged to produce an unnatural completion to an already-begun word.}
All such subwords are penalized by subtracting ${\alpha}n^\beta$ from the output log probabilities of the NMT model before selecting candidates to extend the beam, where $n$ is the n-gram length,
$\alpha$ is the user-specified trade-off between semantic similarity and lexical diversity,  
and $\beta$ is another user-defined hyperparameter.

We experimented with penalizing 1-, 2-, 3-, and 4-grams equally but found it produced disfluent output, as the algorithm tended to avoid all words in the input. The exponential weight allows us to penalize the decoder for producing larger overlapping n-grams more harshly than small ones. All experiments in this work use $\beta=4$, as this produced output in English which appeared fluent to the authors.
Finally, 
the NMT model's vocabulary contains case variants (e.g., ``his'' and ``His'')
and 
we do not want to add variation by trivially changing the case of words, so we penalize all case variants of the next tokens.
Pseudocode for our approach is provided in Algorithm~\ref{algo}.
Note that this method is much simpler than the method used to generate training data for ParaBank 2, which including hand-written constraints, scoring, filtering, and clustering. 

\subsection{Diversity Control}

The $\alpha$ parameter in Equation~\ref{paraphrase-eq} provides the user with a knob to control how strongly the output is ``pushed'' away from the input, in lexical space, during generation.
In contrast to positive and negative hard lexical lexical constraints \cite{hokamp-liu-2017-lexically, post-vilar-2018-fast, hu-etal-2019-large},
our method requires no user-defined constraints, making it simpler and perhaps more language agnostic.\footnote{
One concern with hard constraints is that there are sometimes words or phrases (e.g., proper nouns) that should not be paraphrased, as doing so would change the meaning of the sentence. Thus heuristics are often used to determine which words/phrases should be constrained.
}

\subsection{Development and Evaluation}

Paraphrase evaluation is complicated by the fact that
many different aspects of paraphrases 
can be evaluated including 
semantic similarity between input and output, fluency, grammatical correctness, lexical diversity between input and output, and syntactic diversity between input and output.
The relative importance of these aspects is not intuitively obvious and is likely determined by downstream tasks.

Modeling semantic similarity and lexical/syntactic diversity separately has the potential to somewhat lessen the burden of evaluation in several ways:
\begin{enumerate}[topsep=2pt,itemsep=2pt,partopsep=1pt,parsep=1pt]
\item There are several potential ways
to automatically evaluate the model  $p(y\:{\mid}\:\embed(x))$.
One option is to evaluate perplexity on a test set consisting of human paraphrases.
(\citet{prism} found that their multilingual NMT model assigned higher probability to both copies of the input and human paraphrases of the input, compared to a model trained on ParaBank 2.)
Another option is to test models of $p(y\:{\mid}\:\embed(x))$ on pairs of paraphrases where one paraphrase has been deemed to better preserve the semantic meaning of the input.
Such datasets already exist, in about a dozen languages, due to the annotation efforts undertaken at the annual WMT evaluations.\footnote{In particular, the relative ranking judgements collected through 2016 \cite{bojar-etal-2016-results} are probably the most relevant.}
In other words, we can simply treat a model of $p(y\:{\mid}\:\embed(x))$ as an MT metric in order to judge its quality. 
In other words, we can simply treat a model of $p(y\:{\mid}\:\embed(x))$ as an MT metric in order to judge its quality. 
\item By applying the lexical/syntactic bias in generation,
development of the generation algorithm can be conducted without the time/cost of re-training a model,
and multiple generation schemes can be directly compared using the same $p(y\:{\mid}\:\embed(x))$ model, such as the freely available Prism model \cite{prism}. 
\item Being able to control the amount of lexical and/or syntactic diversity at inference time allows for easier comparison with prior paraphrasing work, as the diversity can be adjusted to match that of a prior method. (We employ this approach in \autoref{fubar}.) 
\end{enumerate}

\section{Experimental Setup}
\label{section:training}

\begin{figure*}
\small
\begin{tabular}{ l|l}
\toprule
Reference         & Among other things, the developments in terms of turnover, employment, warehousing and prices are recorded. \\
\midrule
$\alpha{=}0.0005$ & Among other things, developments in terms of turnover, employment, storage and prices are recorded.         \\
$\alpha{=}0.003$  & Among other things, it records developments in turnover, employment, storage and prices.                    \\
$\alpha{=}0.006$  & Amongst other things, developments regarding turnover, employment, storage and prices were recorded.        \\
\bottomrule
\end{tabular}\caption{Example English paraphrase for the three $\alpha$ values used in this work.}\label{paraphrase-example}
\end{figure*}

\subsection{Primary Model}

We use the multilingual NMT model released with Prism \cite{prism}, which uses a Transformer \cite{vaswani2017attention}
architecture with approximately 750 million parameters.
The model was trained in fairseq \cite{ott2019fairseq}.
The authors take several steps to encourage the encoder and decoder to be language agnostic, including specifying the target language as the first token in the target, so that the encoder does not know the target language, and training on several datasets that include a large number of different language pairs. 
The model was trained on several open source datasets including WikiMatrix
\cite{DBLP:journals/corr/abs-1907-05791},
Global Voices,%
\footnote{\url{http://casmacat.eu/corpus/global-voices.html}}
EuroParl \cite{koehn2005europarl}
SETimes,\footnote{\url{http://nlp.ffzg.hr/resources/corpora/setimes/}} and
 United Nations.
After filtering, this resulted in
approximately 100 million translation pairs
and covering 39 languages.
The model uses a shared, multilingual vocabulary of 64k SentencePiece tokens \cite{kudo-richardson-2018-sentencepiece}.

\subsection{Baseline Model}

As a baseline, we train an English-only paraphraser in fairseq on the ParaBank 2 dataset 
\cite{hu-etal-2019-large} with approximately 253M parameters and a SentencePiece vocabulary of 16k tokens. 
We train a Transformer with an 8-layer encoder, 8-layer decoder,
$1024$ dimensional embeddings, embedding sizes of $1024$,
feed-forward size of $4096$,
and $16$ attention heads.
Dropout is set to $0.3$, label smoothing to $0.1$, and learning rate to $0.0005$, and batch size was 31200 tokens.
Other parameters match the fairseq defaults. 
The model trained for approximately 6 weeks (33 epochs) on 4 Nvidia 2080 GPUs. 

\subsection{Evaluation}

We conduct a manual evaluation in English using Mechanical Turk workers
and conduct smaller scale manual evaluations in German and Spanish, with the help of colleagues who are native speakers.
We perform human evaluations following \cite{hu-etal-2019-parabank}, described in more detail below. 

\subsubsection{English Evaluation}
\label{section:english}\label{fubar}

In this work, we focus on evaluation of semantic similarity, grammatical correctness, and lexical diversity. 
For the model trained on ParaBank 2, 
the trade-off between
these dimensions
is fixed and built into the model. 
To make a fair comparison, 
we adjust our overlap penalty ($\alpha$) 
such that the output of our method matches the lexical diversity of the model trained on ParaBank 2.
Following \citet{hu-etal-2019-large}, 
we use uncased BLEU \cite{papineni-etal-2002-bleu}, 
computed between input and output, 
to estimate the lexical diversity of the paraphraser. %

We evaluate in English using Mechanical Turk workers who were selected 
from a curated list of previously vetted workers.
Annotators were presented with a reference sentence and four paraphrases: 
three paraphrases from our proposed method (at three different operating points) 
and one from the model trained on ParaBank 2, presented in random order. 
For each paraphrase, the annotators were asked to 
(1) rate the paraphrase as (i) grammatical, (ii) having one or two small grammatical errors, or (iii) ungrammatical, and 
(2) rate the semantic similarity between the input and the paraphrase using an analog slider bar from 1--100. 
We randomly select 200 sentences from the English side of the WMT19 German--English test set \citep{barrault-etal-2019-findings} and obtain ratings from three annotators, for each sentence at each paraphrase system/setting combination.
Annotators were paid 0.50 USD per HIT.

For our proposed method, we choose three operating points: 
$\alpha=0.0005$, $\alpha{=}0.003$, and $\alpha{=}0.006$ (\autoref{paraphrase-example}).
The middle point of $\alpha{=}0.003$ was chosen so as to produce output with the same lexical diversity as the paraphraser trained on ParaBank 2, as described above.
We decode with a beam size of 5, using the fairseq defaults.

\subsubsection{German \& Spanish Evaluation}
\label{section:foreign}

We also collect human judgments in German and Spanish. 
We follow the evaluation procedure described above for the English paraphraser except that
annotations were done by colleagues who were native speakers in these languages.
For Spanish, we used the target side of the WMT~2013 English--Spanish test set \citep{bojar-etal-2013-findings}.
For German, we used the target side of the WMT~2019 English--German test set \cite{barrault-etal-2019-findings}.
We obtained 50 judgments per set of 3 paraphrases by one German annotator, and 150 judgments per set of 3 paraphrases by three Spanish annotators, both on a random sample of sentences.
Multiple paraphrases from our proposed method at different operating points (i.e., different values of $\alpha$) were shown to the annotator, in random order.

\section{Results}

\pgfplotsset{compat=1.12}
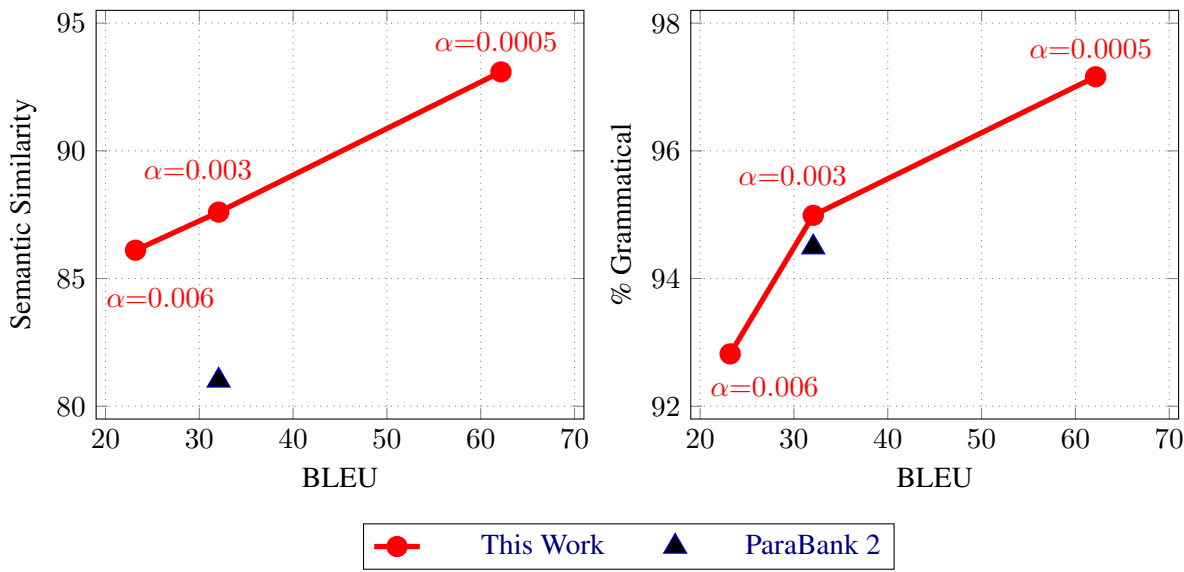
\begin{figure*}[ht]
\centering
  \begin{tikzpicture}
    \begin{groupplot}[group style = {group size = 2 by 1, horizontal sep = 40pt}, 
    width = 8.0cm,
    grid=major,
    xmin=19,
    xmax=71,
major grid style={dotted,line width=.4pt,draw=gray},
    height = 7.0cm]
        \nextgroupplot[  xlabel=BLEU, ylabel=Semantic Similarity,  %
        ymin=79.5,
        ymax=95.5,
            legend style = { column sep = 20pt, legend columns = -1, legend to name = grouplegend,}]
    \addplot [red,
mark size=3pt, mark=*, solid, line width=2pt,mark options={solid,fill=red}] coordinates { (62.16, 93.09) (32.06, 87.60) (23.20, 86.11) };
    \node at (axis cs:54,93.5) [anchor=south west, color=red] {$\alpha{=}0.0005$};
    \node at (axis cs:23,88.5) [anchor=south west, color=red] {$\alpha{=}0.003$};
    \node at (axis cs:19,83.5) [anchor=south west, color=red] {$\alpha{=}0.006$};
    \addlegendentry{This Work}
    \addplot [blue, 
              mark size=5pt, 
              mark=triangle*, only marks, mark options={solid,fill=black},
    ] coordinates { (32.06, 81.01)  };
    \addlegendentry{ParaBank 2}
        \nextgroupplot[ xlabel=BLEU,ylabel=\% Grammatical,
        ymin=91.8,
        ymax=98.2, ]
    \addplot [red, mark size=3pt, mark=*, solid, line width=2pt, mark options={solid,fill=red}] coordinates { (62.16, 97.16) (32.06, 94.99) (23.20, 92.82) };
    \addplot [blue, mark size=5pt, mark=triangle*, only marks, mark options={solid,fill=black}] coordinates { (32.06, 94.49)  };
    \node at (axis cs:54,97.3) [anchor=south west, color=red] {$\alpha{=}0.0005$};
    \node at (axis cs:23,95.3) [anchor=south west, color=red] {$\alpha{=}0.003$};
    \node at (axis cs:20,92) [anchor=south west, color=red] {$\alpha{=}0.006$};
   \end{groupplot}
    \node at ($(group c1r1) + (3.8cm,-4.4cm)$) {\ref{grouplegend}}; 
\end{tikzpicture}

  \caption{Human judgments of English paraphrases for semantic similarity (rated 1--100) and the percentage of sentences produced which were rated as grammatical, both as a function of lexical/syntactic diversity (measured via uncased BLEU between input and output). We evaluated our generation method at three operating points ($\alpha{=}0.0005$, $\alpha{=}0.003$, and $\alpha{=}0.006$). $\alpha{=}0.003$  was chosen to match such that the proposed method had the same diversity as the model trained on Paracrawl2. At that operating point, humans rated output of our method to be more semantically similar to the reference (87.5 vs.\ 81.0), and grammatical slightly more often (95.0\% vs.\ 94.5\%). }\label{human-en-plot}
\end{figure*}

\pgfplotsset{compat=1.12}
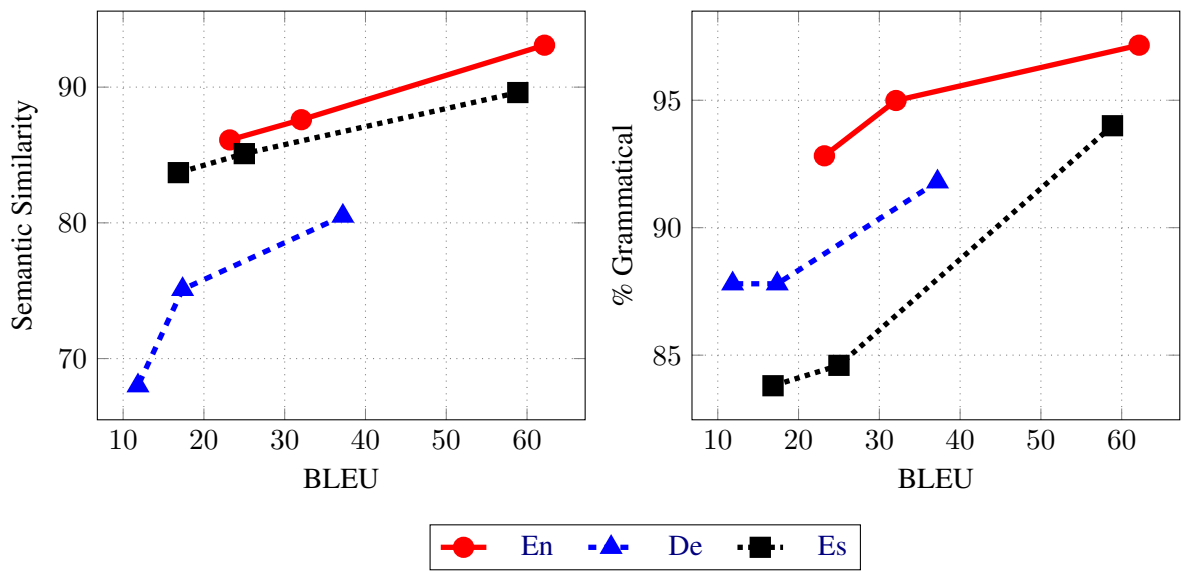
\begin{figure*}[ht!]
\centering
  \begin{tikzpicture}
    \begin{groupplot}[group style = {group size = 2 by 1, horizontal sep = 40pt}, 
    width = 8.0cm,
    grid=major,
    major grid style={dotted,line width=.4pt,draw=gray},
    height = 7.0cm]
    \nextgroupplot[ ylabel = Semantic Similarity, xlabel=BLEU,
    legend style = { column sep = 10pt, legend columns = -1, legend to name = grouplegend,}]
    \addplot [red, mark size=3pt, mark=*, solid, line width=2pt, mark options={solid,fill=red}] coordinates { (62.16, 93.09) (32.06, 87.60) (23.20, 86.11) };
    \addlegendentry{En}
    \addplot [blue, mark size=3pt, mark=triangle*, dashed, line width=2pt, mark options={solid,fill=blue}] 
    coordinates { (37.20, 80.5) (17.36, 75.1) (11.84, 68.0) };
    \addlegendentry{De}
    
    coordinates { (43.23, 50) (20.43, 50) (14.15, 50) };
    \addplot [black, mark size=3pt, mark=square*, dotted, line width=2pt, mark options={solid,fill=black}]
    coordinates { (58.86, 89.6) (25.00, 85.1) (16.84, 83.7) };
    \addlegendentry{Es}
    
    https://www.overleaf.com/project/5db88400f3e1590001826955
    
    \nextgroupplot[ylabel = \% Grammatical,  xlabel=BLEU,]
    \addplot [red, mark size=3pt, mark=*, line width=2pt, solid] coordinates { (62.16, 97.16) (32.06, 94.99) (23.20, 92.82) };
    \addplot [blue, mark size=3pt, mark=triangle*, dashed, line width=2pt, mark options={solid,fill=blue}] 
    coordinates { (37.20, 91.8) (17.36, 87.8) (11.84, 87.8) };
    \addplot [black, mark size=3pt, mark=square*, dotted, line width=2pt, mark options={solid,fill=black}] 
    coordinates { (58.86, 94.0) (25.00, 84.6) (16.84, 83.8) };

    \end{groupplot}
    \node at ($(group c1r1) + (4.0cm,-4.4cm)$) {\ref{grouplegend}}; 
\end{tikzpicture}

  \caption{Human judgments of German (De) and Spanish (Es) paraphrases, with English (En) shown for reference, plotted against uncased BLEU computed between the paraphraser input and output. 
  The judgement criteria and $\alpha$ values match English settings.
  $\alpha$ decreases from left to right in all plots. 
  }\label{human-fl-plot}
\end{figure*}

\subsection{English Results}

Human evaluation results in English are shown in \autoref{human-en-plot}.
We find that $\alpha$
is negatively correlated with grammaticality and semantic similarity between the input and output
and positively correlated with lexical diversity of the output with respect to the input, as expected. 

We find that at the operating point $\alpha = 0.003$, which was chosen such that our method has the same lexical diversity as the model trained on ParaBank 2, the paraphrases from our method were judged to be both more semantically similar to the input and grammatical (slightly) more often.

\subsection{German \& Spanish Results}

The human evaluation results in German and Spanish, along with English for reference, are shown in \autoref{human-fl-plot}. 
Note that we have no way to normalize between annotators in different languages, thus the results should \emph{not} be used to draw conclusions about the \emph{relative} performance of the paraphraser of these languages.
However, we find the trends are similar across all three languages, and that semantic similarity and grammaticality judgements for Spanish and German are both reasonably high.

\section{Discussion}\label{discussion}

We hypothesize that our method outperforms the baseline because it does not suffer from a fundamental 
shortcoming in creating synthetic paraphrase data from bitext: 
namely that inherent ambiguities present in one language (but not the other) 
can cause erroneous synthetic paraphrases in the other language \cite{aziz-specia-2013-multilingual}.

For the sake of discussion we consider 
gender\footnote{Czech is, of course, gendered, 
so we would not expect the ParaBank 2 dataset 
(which was created from Czech--English bitext) to have gender errors. 
But the logic presented here should generalize to other ambiguities.} as an ambiguity. 
Suppose we create synthetic English paraphrases from Turkish--English data, 
and our bitext contains the following (valid) sentence pair: 
(``O mağazaya gitti.'', ``She went to the store.'')
Turkish is a gender-neutral language, 
so when we translate the Turkish side to English it is perfectly valid 
to translate the sentence to ``He went to the store.'' 
Pairing the original English translation with the translation results 
in the synthetic paraphrase example (``She went to the store.'', ``He went to the store.''). 
Since English is gendered, this results in an invalid synthetic paraphrase.

In contrast, consider what happens if ``She went to the store.'' 
is paraphrased by our method. 
First, the sentence is converted to an intermediate representation by the encoder.
If the encoder were from an English$\rightarrow$Turkish system, 
it is plausible that the encoder would discard gender information, 
as it is not needed in the target language. 
However, our encoder comes from a multilingual system 
which can produce output in \emph{many} different languages. 
Thus, as long as the model has seen a sufficient number of training examples 
between English and at least one other gendered language, 
we can reasonably expect that the intermediate representation will preserve gender. 
Thus, when this representation is passed to the decoder and English is requested as the target language, 
the model should put low probability on any output for which the subject is male.

An alternative way to address pivot language ambiguities is to use multiple pivot languages,
as proposed by \citet{aziz-specia-2013-multilingual}.
However, it is not clear how best to extend this idea to neural sequence-to-sequence models, or to a multilingual paraphraser.
Combining synthetic paraphrases for training using several different pivot languages would mitigate the errors due
to ambiguities from any one pivot language, at the
expense of errors due to ambiguities in other pivot
languages. To really address such errors would
require combining models of different language pairs;
see \citet{mallinson-etal-2017-paraphrasing} for one such solution.

\section{Conclusions}

We treat paraphrasing as a zero-shot translation task
and 
present a method to control the lexical diversity 
of paraphrases generated from a multilingual NMT model, 
enabling paraphrase generation in many languages. 
Our approach gives a user fine-grained control over the amount of
lexical diversity at generation time,
and also allows models and generation algorithms to be developed and evaluated with less interdependencies. 
There are likely many other ways that the output could be 
controlled to vary other aspects, 
such as syntactic diversity \cite{shu-etal-2019-generating};
we would like to explore such methods in future work.

Our work outperforms an English baseline trained 
on a large synthetic paraphrase dataset \cite{hu-etal-2019-parabank}.
This improvement in performance may be because our method 
does not suffer from the issue that ambiguities in the pivot language used 
to create synthetic paraphrase data can cause errors in synthetic data. 
Small experiments indicate our method also performs well in other languages.

Multilingual NMT is an active research area and 
we are optimistic that this approach will pave the way 
for even stronger paraphrase generation in the future, 
as multilingual NMT methods continue to improve
and models are publicly released.

\section*{Acknowledgments}
The authors wish to thank Ben Van Durme for helpful technical discussions, and Carlos Aguirre, Sabrina Mielke, Rachel Wicks, and others for assistance with annotations.

\clearpage   %

\bibliographystyle{acl_natbib}
\bibliography{anthology,emnlp2020}

\begin{thebibliography}{46}
\expandafter\ifx\csname natexlab\endcsname\relax\def\natexlab#1{#1}\fi

\bibitem[{Aziz and Specia(2013)}]{aziz-specia-2013-multilingual}
Wilker Aziz and Lucia Specia. 2013.
\newblock \href {https://www.aclweb.org/anthology/W13-3522} {Multilingual
  {WSD}-like constraints for paraphrase extraction}.
\newblock In \emph{Proceedings of the Seventeenth Conference on Computational
  Natural Language Learning}, pages 202--211, Sofia, Bulgaria. Association for
  Computational Linguistics.

\bibitem[{Banerjee and Lavie(2005)}]{banerjee-lavie-2005-meteor}
Satanjeev Banerjee and Alon Lavie. 2005.
\newblock \href {https://www.aclweb.org/anthology/W05-0909} {{METEOR}: An
  automatic metric for {MT} evaluation with improved correlation with human
  judgments}.
\newblock In \emph{Proceedings of the {ACL} Workshop on Intrinsic and Extrinsic
  Evaluation Measures for Machine Translation and/or Summarization}, pages
  65--72, Ann Arbor, Michigan. Association for Computational Linguistics.

\bibitem[{Barrault et~al.(2019)Barrault, Bojar, Costa-juss{\`a}, Federmann,
  Fishel, Graham, Haddow, Huck, Koehn, Malmasi, Monz, M{\"u}ller, Pal, Post,
  and Zampieri}]{barrault-etal-2019-findings}
Lo{\"\i}c Barrault, Ond{\v{r}}ej Bojar, Marta~R. Costa-juss{\`a}, Christian
  Federmann, Mark Fishel, Yvette Graham, Barry Haddow, Matthias Huck, Philipp
  Koehn, Shervin Malmasi, Christof Monz, Mathias M{\"u}ller, Santanu Pal, Matt
  Post, and Marcos Zampieri. 2019.
\newblock \href {https://doi.org/10.18653/v1/W19-5301} {Findings of the 2019
  conference on machine translation ({WMT}19)}.
\newblock In \emph{Proceedings of the Fourth Conference on Machine Translation
  (Volume 2: Shared Task Papers, Day 1)}, pages 1--61, Florence, Italy.
  Association for Computational Linguistics.

\bibitem[{Bhagat and Hovy(2013)}]{bhagat-hovy-2013-squibs}
Rahul Bhagat and Eduard Hovy. 2013.
\newblock \href {https://doi.org/10.1162/COLI_a_00166} {{S}quibs: What is a
  paraphrase?}
\newblock \emph{Computational Linguistics}, 39(3):463--472.

\bibitem[{Bojar et~al.(2013)Bojar, Buck, Callison-Burch, Federmann, Haddow,
  Koehn, Monz, Post, Soricut, and Specia}]{bojar-etal-2013-findings}
Ond{\v{r}}ej Bojar, Christian Buck, Chris Callison-Burch, Christian Federmann,
  Barry Haddow, Philipp Koehn, Christof Monz, Matt Post, Radu Soricut, and
  Lucia Specia. 2013.
\newblock \href {https://www.aclweb.org/anthology/W13-2201} {Findings of the
  2013 {W}orkshop on {S}tatistical {M}achine {T}ranslation}.
\newblock In \emph{Proceedings of the Eighth Workshop on Statistical Machine
  Translation}, pages 1--44, Sofia, Bulgaria. Association for Computational
  Linguistics.

\bibitem[{Bojar et~al.(2016)Bojar, Graham, Kamran, and
  Stanojevi{\'c}}]{bojar-etal-2016-results}
Ond{\v{r}}ej Bojar, Yvette Graham, Amir Kamran, and Milo{\v{s}} Stanojevi{\'c}.
  2016.
\newblock \href {https://doi.org/10.18653/v1/W16-2302} {Results of the {WMT}16
  metrics shared task}.
\newblock In \emph{Proceedings of the First Conference on Machine Translation:
  Volume 2, Shared Task Papers}, pages 199--231, Berlin, Germany. Association
  for Computational Linguistics.

\bibitem[{Chen et~al.(2019{\natexlab{a}})Chen, Tang, Wiseman, and
  Gimpel}]{chen-etal-2019-controllable}
Mingda Chen, Qingming Tang, Sam Wiseman, and Kevin Gimpel. 2019{\natexlab{a}}.
\newblock \href {https://doi.org/10.18653/v1/P19-1599} {Controllable paraphrase
  generation with a syntactic exemplar}.
\newblock In \emph{Proceedings of the 57th Annual Meeting of the Association
  for Computational Linguistics}, pages 5972--5984, Florence, Italy.
  Association for Computational Linguistics.

\bibitem[{Chen et~al.(2019{\natexlab{b}})Chen, Tang, Wiseman, and
  Gimpel}]{chen-etal-2019-multi}
Mingda Chen, Qingming Tang, Sam Wiseman, and Kevin Gimpel. 2019{\natexlab{b}}.
\newblock \href {https://doi.org/10.18653/v1/N19-1254} {A multi-task approach
  for disentangling syntax and semantics in sentence representations}.
\newblock In \emph{Proceedings of the 2019 Conference of the North {A}merican
  Chapter of the Association for Computational Linguistics: Human Language
  Technologies, Volume 1 (Long and Short Papers)}, pages 2453--2464,
  Minneapolis, Minnesota. Association for Computational Linguistics.

\bibitem[{Denkowski and Lavie(2010)}]{denkowski-lavie-2010-extending}
Michael Denkowski and Alon Lavie. 2010.
\newblock \href {https://www.aclweb.org/anthology/N10-1031} {Extending the
  {METEOR} machine translation evaluation metric to the phrase level}.
\newblock In \emph{Human Language Technologies: The 2010 Annual Conference of
  the North {A}merican Chapter of the Association for Computational
  Linguistics}, pages 250--253, Los Angeles, California. Association for
  Computational Linguistics.

\bibitem[{Dong et~al.(2015)Dong, Wu, He, Yu, and Wang}]{dong-etal-2015-multi}
Daxiang Dong, Hua Wu, Wei He, Dianhai Yu, and Haifeng Wang. 2015.
\newblock \href {https://doi.org/10.3115/v1/P15-1166} {Multi-task learning for
  multiple language translation}.
\newblock In \emph{Proceedings of the 53rd Annual Meeting of the Association
  for Computational Linguistics and the 7th International Joint Conference on
  Natural Language Processing (Volume 1: Long Papers)}, pages 1723--1732,
  Beijing, China. Association for Computational Linguistics.

\bibitem[{Dong et~al.(2017)Dong, Mallinson, Reddy, and
  Lapata}]{dong-etal-2017-learning}
Li~Dong, Jonathan Mallinson, Siva Reddy, and Mirella Lapata. 2017.
\newblock \href {https://doi.org/10.18653/v1/D17-1091} {Learning to paraphrase
  for question answering}.
\newblock In \emph{Proceedings of the 2017 Conference on Empirical Methods in
  Natural Language Processing}, pages 875--886, Copenhagen, Denmark.
  Association for Computational Linguistics.

\bibitem[{Gan and Ng(2019)}]{gan-ng-2019-improving}
Wee~Chung Gan and Hwee~Tou Ng. 2019.
\newblock \href {https://doi.org/10.18653/v1/P19-1610} {Improving the
  robustness of question answering systems to question paraphrasing}.
\newblock In \emph{Proceedings of the 57th Annual Meeting of the Association
  for Computational Linguistics}, pages 6065--6075, Florence, Italy.
  Association for Computational Linguistics.

\bibitem[{Gu et~al.(2018)Gu, Hassan, Devlin, and Li}]{gu-etal-2018-universal}
Jiatao Gu, Hany Hassan, Jacob Devlin, and Victor~O.K. Li. 2018.
\newblock \href {https://doi.org/10.18653/v1/N18-1032} {Universal neural
  machine translation for extremely low resource languages}.
\newblock In \emph{Proceedings of the 2018 Conference of the North {A}merican
  Chapter of the Association for Computational Linguistics: Human Language
  Technologies, Volume 1 (Long Papers)}, pages 344--354, New Orleans,
  Louisiana. Association for Computational Linguistics.

\bibitem[{Gupta et~al.(2018)Gupta, Agarwal, Singh, and Rai}]{AAAI1816353}
Ankush Gupta, Arvind Agarwal, Prawaan Singh, and Piyush Rai. 2018.
\newblock A deep generative framework for paraphrase generation.
\newblock In \emph{AAAI Conference on Artificial Intelligence}.

\bibitem[{Hokamp and Liu(2017)}]{hokamp-liu-2017-lexically}
Chris Hokamp and Qun Liu. 2017.
\newblock \href {https://doi.org/10.18653/v1/P17-1141} {Lexically constrained
  decoding for sequence generation using grid beam search}.
\newblock In \emph{Proceedings of the 55th Annual Meeting of the Association
  for Computational Linguistics (Volume 1: Long Papers)}, pages 1535--1546,
  Vancouver, Canada. Association for Computational Linguistics.

\bibitem[{Hu et~al.(2019{\natexlab{a}})Hu, Khayrallah, Culkin, Xia, Chen, Post,
  and Van~Durme}]{hu-etal-2019-improved}
J.~Edward Hu, Huda Khayrallah, Ryan Culkin, Patrick Xia, Tongfei Chen, Matt
  Post, and Benjamin Van~Durme. 2019{\natexlab{a}}.
\newblock \href {https://doi.org/10.18653/v1/N19-1090} {Improved lexically
  constrained decoding for translation and monolingual rewriting}.
\newblock In \emph{Proceedings of the 2019 Conference of the North {A}merican
  Chapter of the Association for Computational Linguistics: Human Language
  Technologies, Volume 1 (Long and Short Papers)}, pages 839--850, Minneapolis,
  Minnesota. Association for Computational Linguistics.

\bibitem[{Hu et~al.(2019{\natexlab{b}})Hu, Rudinger, Post, and {Van
  Durme}}]{hu-etal-2019-parabank}
J.~Edward Hu, Rachel Rudinger, Matt Post, and Benjamin {Van Durme}.
  2019{\natexlab{b}}.
\newblock \href {https://arxiv.org/pdf/1901.03644.pdf} {Para{B}ank: Monolingual
  bitext generation and sentential paraphrasing via lexically-constrained
  neural machine translation}.
\newblock In \emph{Proceedings of AAAI}.

\bibitem[{Hu et~al.(2019{\natexlab{c}})Hu, Singh, Holzenberger, Post, and
  Van~Durme}]{hu-etal-2019-large}
J.~Edward Hu, Abhinav Singh, Nils Holzenberger, Matt Post, and Benjamin
  Van~Durme. 2019{\natexlab{c}}.
\newblock \href {https://doi.org/10.18653/v1/K19-1005} {Large-scale, diverse,
  paraphrastic bitexts via sampling and clustering}.
\newblock In \emph{Proceedings of the 23rd Conference on Computational Natural
  Language Learning (CoNLL)}, pages 44--54, Hong Kong, China. Association for
  Computational Linguistics.

\bibitem[{Iyyer et~al.(2018)Iyyer, Wieting, Gimpel, and
  Zettlemoyer}]{iyyer-etal-2018-adversarial}
Mohit Iyyer, John Wieting, Kevin Gimpel, and Luke Zettlemoyer. 2018.
\newblock \href {https://doi.org/10.18653/v1/N18-1170} {Adversarial example
  generation with syntactically controlled paraphrase networks}.
\newblock In \emph{Proceedings of the 2018 Conference of the North {A}merican
  Chapter of the Association for Computational Linguistics: Human Language
  Technologies, Volume 1 (Long Papers)}, pages 1875--1885, New Orleans,
  Louisiana. Association for Computational Linguistics.

\bibitem[{Johnson et~al.(2017)Johnson, Schuster, Le, Krikun, Wu, Chen, Thorat,
  Vi{\'e}gas, Wattenberg, Corrado, Hughes, and
  Dean}]{johnson-etal-2017-googles}
Melvin Johnson, Mike Schuster, Quoc~V. Le, Maxim Krikun, Yonghui Wu, Zhifeng
  Chen, Nikhil Thorat, Fernanda Vi{\'e}gas, Martin Wattenberg, Greg Corrado,
  Macduff Hughes, and Jeffrey Dean. 2017.
\newblock \href {https://doi.org/10.1162/tacl_a_00065} {{G}oogle{'}s
  multilingual neural machine translation system: Enabling zero-shot
  translation}.
\newblock \emph{Transactions of the Association for Computational Linguistics},
  5:339--351.

\bibitem[{Kajiwara(2019)}]{kajiwara-2019-negative}
Tomoyuki Kajiwara. 2019.
\newblock \href {https://doi.org/10.18653/v1/P19-1607} {Negative lexically
  constrained decoding for paraphrase generation}.
\newblock In \emph{Proceedings of the 57th Annual Meeting of the Association
  for Computational Linguistics}, pages 6047--6052, Florence, Italy.
  Association for Computational Linguistics.

\bibitem[{Khayrallah et~al.(2020)Khayrallah, Thompson, Post, and
  Koehn}]{khayrallah-etal-2020-simulated}
Huda Khayrallah, Brian Thompson, Matt Post, and Philipp Koehn. 2020.
\newblock Simulated multiple reference training improves low-resource machine
  translation.
\newblock In \emph{Proceedings of the 2020 Conference on Empirical Methods in
  Natural Language Processing}, Online. Association for Computational
  Linguistics.

\bibitem[{Koehn(2005)}]{koehn2005europarl}
Philipp Koehn. 2005.
\newblock Europarl: A parallel corpus for statistical machine translation.
\newblock In \emph{MT summit}, volume~5, pages 79--86. Citeseer.

\bibitem[{Kudo and Richardson(2018)}]{kudo-richardson-2018-sentencepiece}
Taku Kudo and John Richardson. 2018.
\newblock \href {https://doi.org/10.18653/v1/D18-2012} {{S}entence{P}iece: A
  simple and language independent subword tokenizer and detokenizer for neural
  text processing}.
\newblock In \emph{Proceedings of the 2018 Conference on Empirical Methods in
  Natural Language Processing: System Demonstrations}, pages 66--71, Brussels,
  Belgium. Association for Computational Linguistics.

\bibitem[{Li et~al.(2018)Li, Jiang, Shang, and Li}]{li-etal-2018-paraphrase}
Zichao Li, Xin Jiang, Lifeng Shang, and Hang Li. 2018.
\newblock \href {https://doi.org/10.18653/v1/D18-1421} {Paraphrase generation
  with deep reinforcement learning}.
\newblock In \emph{Proceedings of the 2018 Conference on Empirical Methods in
  Natural Language Processing}, pages 3865--3878, Brussels, Belgium.
  Association for Computational Linguistics.

\bibitem[{Mallinson et~al.(2017)Mallinson, Sennrich, and
  Lapata}]{mallinson-etal-2017-paraphrasing}
Jonathan Mallinson, Rico Sennrich, and Mirella Lapata. 2017.
\newblock \href {https://www.aclweb.org/anthology/E17-1083} {Paraphrasing
  revisited with neural machine translation}.
\newblock In \emph{Proceedings of the 15th Conference of the {E}uropean Chapter
  of the Association for Computational Linguistics: Volume 1, Long Papers},
  pages 881--893, Valencia, Spain. Association for Computational Linguistics.

\bibitem[{McKeown(1983)}]{mckeown-1983-paraphrasing}
Kathleen~R. McKeown. 1983.
\newblock \href {https://www.aclweb.org/anthology/J83-1001} {Paraphrasing
  questions using given and new information}.
\newblock \emph{American Journal of Computational Linguistics}, 9(1):1--10.

\bibitem[{Niu and Bansal(2018)}]{niu-bansal-2018-adversarial}
Tong Niu and Mohit Bansal. 2018.
\newblock \href {https://doi.org/10.18653/v1/K18-1047} {Adversarial
  over-sensitivity and over-stability strategies for dialogue models}.
\newblock In \emph{Proceedings of the 22nd Conference on Computational Natural
  Language Learning}, pages 486--496, Brussels, Belgium. Association for
  Computational Linguistics.

\bibitem[{Niu and Bansal(2019)}]{niu-bansal-2019-automatically}
Tong Niu and Mohit Bansal. 2019.
\newblock \href {https://doi.org/10.18653/v1/D19-1132} {Automatically learning
  data augmentation policies for dialogue tasks}.
\newblock In \emph{Proceedings of the 2019 Conference on Empirical Methods in
  Natural Language Processing and the 9th International Joint Conference on
  Natural Language Processing (EMNLP-IJCNLP)}, pages 1317--1323, Hong Kong,
  China. Association for Computational Linguistics.

\bibitem[{Ott et~al.(2019)Ott, Edunov, Baevski, Fan, Gross, Ng, Grangier, and
  Auli}]{ott2019fairseq}
Myle Ott, Sergey Edunov, Alexei Baevski, Angela Fan, Sam Gross, Nathan Ng,
  David Grangier, and Michael Auli. 2019.
\newblock \href {https://doi.org/10.18653/v1/N19-4009} {fairseq: A fast,
  extensible toolkit for sequence modeling}.
\newblock In \emph{Proceedings of the 2019 Conference of the North {A}merican
  Chapter of the Association for Computational Linguistics (Demonstrations)},
  pages 48--53, Minneapolis, Minnesota. Association for Computational
  Linguistics.

\bibitem[{Papineni et~al.(2002)Papineni, Roukos, Ward, and
  Zhu}]{papineni-etal-2002-bleu}
Kishore Papineni, Salim Roukos, Todd Ward, and Wei-Jing Zhu. 2002.
\newblock \href {https://doi.org/10.3115/1073083.1073135} {{B}leu: a method for
  automatic evaluation of machine translation}.
\newblock In \emph{Proceedings of the 40th Annual Meeting of the Association
  for Computational Linguistics}, pages 311--318, Philadelphia, Pennsylvania,
  USA. Association for Computational Linguistics.

\bibitem[{Pham et~al.(2019)Pham, Niehues, Ha, and
  Waibel}]{pham-etal-2019-improving}
Ngoc-Quan Pham, Jan Niehues, Thanh-Le Ha, and Alexander Waibel. 2019.
\newblock \href {https://doi.org/10.18653/v1/W19-5202} {Improving zero-shot
  translation with language-independent constraints}.
\newblock In \emph{Proceedings of the Fourth Conference on Machine Translation
  (Volume 1: Research Papers)}, pages 13--23, Florence, Italy. Association for
  Computational Linguistics.

\bibitem[{Post and Vilar(2018)}]{post-vilar-2018-fast}
Matt Post and David Vilar. 2018.
\newblock \href {https://doi.org/10.18653/v1/N18-1119} {Fast lexically
  constrained decoding with dynamic beam allocation for neural machine
  translation}.
\newblock In \emph{Proceedings of the 2018 Conference of the North {A}merican
  Chapter of the Association for Computational Linguistics: Human Language
  Technologies, Volume 1 (Long Papers)}, pages 1314--1324, New Orleans,
  Louisiana. Association for Computational Linguistics.

\bibitem[{Quirk et~al.(2004)Quirk, Brockett, and
  Dolan}]{quirk-etal-2004-monolingual}
Chris Quirk, Chris Brockett, and William Dolan. 2004.
\newblock \href {https://www.aclweb.org/anthology/W04-3219} {Monolingual
  machine translation for paraphrase generation}.
\newblock In \emph{Proceedings of the 2004 Conference on Empirical Methods in
  Natural Language Processing}, pages 142--149, Barcelona, Spain. Association
  for Computational Linguistics.

\bibitem[{Raganato et~al.(2019)Raganato, V{\'a}zquez, Creutz, and
  Tiedemann}]{raganato-etal-2019-evaluation}
Alessandro Raganato, Ra{\'u}l V{\'a}zquez, Mathias Creutz, and J{\"o}rg
  Tiedemann. 2019.
\newblock \href {https://doi.org/10.18653/v1/W19-4304} {An evaluation of
  language-agnostic inner-attention-based representations in machine
  translation}.
\newblock In \emph{Proceedings of the 4th Workshop on Representation Learning
  for NLP (RepL4NLP-2019)}, pages 27--32, Florence, Italy. Association for
  Computational Linguistics.

\bibitem[{Schwenk et~al.(2019)Schwenk, Chaudhary, Sun, Gong, and
  Guzm{\'{a}}n}]{DBLP:journals/corr/abs-1907-05791}
Holger Schwenk, Vishrav Chaudhary, Shuo Sun, Hongyu Gong, and Francisco
  Guzm{\'{a}}n. 2019.
\newblock \href {http://arxiv.org/abs/1907.05791} {Wikimatrix: Mining 135m
  parallel sentences in 1620 language pairs from wikipedia}.
\newblock \emph{CoRR}, abs/1907.05791.

\bibitem[{Schwenk and Douze(2017)}]{schwenk-douze-2017-learning}
Holger Schwenk and Matthijs Douze. 2017.
\newblock \href {https://doi.org/10.18653/v1/W17-2619} {Learning joint
  multilingual sentence representations with neural machine translation}.
\newblock In \emph{Proceedings of the 2nd Workshop on Representation Learning
  for {NLP}}, pages 157--167, Vancouver, Canada. Association for Computational
  Linguistics.

\bibitem[{Shu et~al.(2019)Shu, Nakayama, and Cho}]{shu-etal-2019-generating}
Raphael Shu, Hideki Nakayama, and Kyunghyun Cho. 2019.
\newblock \href {https://doi.org/10.18653/v1/P19-1177} {Generating diverse
  translations with sentence codes}.
\newblock In \emph{Proceedings of the 57th Annual Meeting of the Association
  for Computational Linguistics}, pages 1823--1827, Florence, Italy.
  Association for Computational Linguistics.

\bibitem[{Thompson and Post(2020)}]{prism}
Brian Thompson and Matt Post. 2020.
\newblock Automatic machine translation evaluation in many languages via
  zero-shot paraphrasing.
\newblock In \emph{Proceedings of the 2020 Conference on Empirical Methods in
  Natural Language Processing}, Online. Association for Computational
  Linguistics.

\bibitem[{Tiedemann and Scherrer(2019)}]{tiedemann-scherrer-2019-measuring}
J{\"o}rg Tiedemann and Yves Scherrer. 2019.
\newblock \href {https://doi.org/10.18653/v1/W19-2005} {Measuring semantic
  abstraction of multilingual {NMT} with paraphrase recognition and generation
  tasks}.
\newblock In \emph{Proceedings of the 3rd Workshop on Evaluating Vector Space
  Representations for {NLP}}, pages 35--42, Minneapolis, USA. Association for
  Computational Linguistics.

\bibitem[{Vaswani et~al.(2017)Vaswani, Shazeer, Parmar, Uszkoreit, Jones,
  Gomez, Kaiser, and Polosukhin}]{vaswani2017attention}
Ashish Vaswani, Noam Shazeer, Niki Parmar, Jakob Uszkoreit, Llion Jones,
  Aidan~N Gomez, \L~ukasz Kaiser, and Illia Polosukhin. 2017.
\newblock \href
  {http://papers.nips.cc/paper/7181-attention-is-all-you-need.pdf} {Attention
  is all you need}.
\newblock In I.~Guyon, U.~V. Luxburg, S.~Bengio, H.~Wallach, R.~Fergus,
  S.~Vishwanathan, and R.~Garnett, editors, \emph{Advances in Neural
  Information Processing Systems 30}, pages 5998--6008. Curran Associates, Inc.

\bibitem[{Wieting and Gimpel(2018)}]{wieting-gimpel-2018-paranmt}
John Wieting and Kevin Gimpel. 2018.
\newblock \href {https://doi.org/10.18653/v1/P18-1042} {{P}ara{NMT}-50{M}:
  Pushing the limits of paraphrastic sentence embeddings with millions of
  machine translations}.
\newblock In \emph{Proceedings of the 56th Annual Meeting of the Association
  for Computational Linguistics (Volume 1: Long Papers)}, pages 451--462,
  Melbourne, Australia. Association for Computational Linguistics.

\bibitem[{Wieting et~al.(2019)Wieting, Gimpel, Neubig, and
  Berg-Kirkpatrick}]{wieting-etal-2019-simple}
John Wieting, Kevin Gimpel, Graham Neubig, and Taylor Berg-Kirkpatrick. 2019.
\newblock \href {https://doi.org/10.18653/v1/P19-1453} {Simple and effective
  paraphrastic similarity from parallel translations}.
\newblock In \emph{Proceedings of the 57th Annual Meeting of the Association
  for Computational Linguistics}, pages 4602--4608, Florence, Italy.
  Association for Computational Linguistics.

\bibitem[{Wieting et~al.(2017)Wieting, Mallinson, and
  Gimpel}]{wieting-etal-2017-learning}
John Wieting, Jonathan Mallinson, and Kevin Gimpel. 2017.
\newblock \href {https://doi.org/10.18653/v1/D17-1026} {Learning paraphrastic
  sentence embeddings from back-translated bitext}.
\newblock In \emph{Proceedings of the 2017 Conference on Empirical Methods in
  Natural Language Processing}, pages 274--285, Copenhagen, Denmark.
  Association for Computational Linguistics.

\bibitem[{Zhou et~al.(2006)Zhou, Lin, Munteanu, and
  Hovy}]{zhou-etal-2006-paraeval}
Liang Zhou, Chin-Yew Lin, Dragos~Stefan Munteanu, and Eduard Hovy. 2006.
\newblock \href {https://www.aclweb.org/anthology/N06-1057} {{P}ara{E}val:
  Using paraphrases to evaluate summaries automatically}.
\newblock In \emph{Proceedings of the Human Language Technology Conference of
  the {NAACL}, Main Conference}, pages 447--454, New York City, USA.
  Association for Computational Linguistics.

\bibitem[{Zhou et~al.(2019)Zhou, Sperber, and
  Waibel}]{zhou-etal-2019-paraphrases}
Zhong Zhou, Matthias Sperber, and Alexander Waibel. 2019.
\newblock \href {https://doi.org/10.18653/v1/P19-2015} {Paraphrases as foreign
  languages in multilingual neural machine translation}.
\newblock In \emph{Proceedings of the 57th Annual Meeting of the Association
  for Computational Linguistics: Student Research Workshop}, pages 113--122,
  Florence, Italy. Association for Computational Linguistics.

\end{thebibliography}

\end{document}